# Interaction valence reveals contrasting social networks in dairy cattle

**Sibi Parivendan[1], and Suresh Raja Neethirajan[1,2*]**
[1]Faculty of Computer Science, Dalhousie University, Halifax, NS, Canada
[2]Faculty of Agriculture, Dalhousie University, Truro, NS, Canada
* Correspondence: sneethir@gmail.com

**Abstract**
Social relationships shape access to resources, exposure to conflict and group stability, yet automated livestock monitoring typically treats behaviour as isolated events. Here, we present a valence-aware social-network framework that transforms video-derived interactions into herd-level representations of affiliative and agonistic organization. A pose-based computer-vision pipeline analysed 7 h 39 min of continuous video from the pre-milking area of one commercial dairy farm. After quality control, 1,183 of 1,414 candidate interactions remained, involving 36 cows and 177 dyads. In a predicted-class-balanced audit of 198 pipeline-detected clips, automated and manual labels agreed in 82.8% of cases, with an unweighted audit-sample macro-F1 of 0.872. These values describe the audited sample rather than prevalence-weighted or end-to-end deployment performance. The aggregated network was connected (density = 0.281; transitivity = 0.513; mean path length = 1.88), and predicted affiliative events formed five algorithmic communities (modularity $Q$ = 0.429). Within the observed zone, predicted agonistic interactions comprised 72.4% of retained events and 76.0% of interaction duration. The cow with the most partners did not have the highest betweenness centrality. Separating events by predicted valence produced descriptively different affiliative and agonistic layers, with contrasting edge sets, community partitions and individual positions. Thus, pooled interaction counts can obscure the behavioural composition of an observed network. Valence-aware analysis provides a framework for testing hypotheses about competition, affiliation and welfare-relevant change, while requiring longitudinal validation before use as a welfare or health indicator.

## 1. Introduction

Precision livestock farming has made dairy cows increasingly legible as individuals. Automated milking systems, precision feeding, wearable sensors, machine vision, and integrated data platforms can now track production, physiology, health, and behaviour at unprecedented resolution [1]. Yet the dominant unit of analysis remains the individual animal. This is a consequential simplification. Cattle live in socially structured groups, and an animal’s access to feed, water, resting space, and shared passageways depends partly on its encounters with herdmates [2, 3]. Herds are therefore not merely collections of monitored individuals; they are networks of affiliation and competition in which animals occupy different positions, form subgroups, and sometimes connect otherwise separated parts of the group. These positions can shape access to resources, exposure to social stress, and responses to disruptions such as regrouping [4, 5]. A monitoring system that resolves individual state while ignoring relational structure captures only part of the biology of the herd.

Social organization remains largely invisible in commercial settings because it is difficult to measure continuously. Manual observation is slow, observer-dependent, and difficult to sustain at herd scale in a working barn [6]. This measurement gap matters most around heavily used resources, where repeated encounters can accumulate into unequal social experience. Agonistic behaviours, including displacement, blocking, and pushing, become more frequent when animals compete for feed, resting space, or passage [4, 7]. Such behaviours are normal components of group living, but sustained competitive pressure may restrict resource access, increase social stress, and distribute its costs unevenly across animals [5, 8]. Affiliative behaviours, including allogrooming and gentle social contact, provide a complementary view of group organization by reinforcing familiarity and cohesion [9, 10]. Consequently, interaction frequency alone is insufficient: a highly connected cow may be socially integrated, repeatedly challenged, or both. Distinguishing the valence of those encounters is central to interpreting their biological meaning [11, 12]. This relational perspective is consistent with the emerging concept of social welfare, which recognizes the social environment as one of the conditions shaping animal well-being [13]. Social networks cannot by themselves diagnose welfare state, but they can characterize aspects of the social context that individual production measures do not capture.

Social network analysis provides a quantitative language for this relational biology [13]. It represents individuals as nodes and their relationships as edges, allowing the organization of a group to be examined through measures of connectivity, brokerage, local cohesion, and community structure. Across animal systems, network analysis has been used to study the movement of information, infection, and social influence and to distinguish the roles occupied by different individuals [1]. Dairy cattle are particularly amenable to this approach because individuals can be identified reliably and routine activities repeatedly bring the same animals together [14–16]. Existing cattle networks have been constructed from manual observations and from proximity or contact data collected using radio-frequency identification, positioning systems, and wearable sensors [17, 18]. These studies have revealed stable social preferences, dominance relationships, and contact structures relevant to disease transmission [5, 14, 19].

Most automated cattle networks nevertheless share a fundamental interpretive limitation: proximity is not interaction. A link based on co-location records an opportunity for social exchange, but not what occurred. Two cows standing together at a feed bunk may be feeding peacefully, competing for access, or merely occupying the same constrained space [14, 18]. The ambiguity propagates into network interpretation. A central animal may be socially integrated, repeatedly involved in conflict, or simply present at a high-traffic location; an apparent community may represent an affiliative subgroup or a set of animals drawn independently to the same resource. Even when observable encounters are counted, pooling all events treats gentle contact and displacement as equivalent evidence of association. Interaction counts alone can therefore obscure social organization: they record how often animals meet while collapsing what those meetings mean. A network may be mathematically well resolved yet behaviourally under-specified. Recovering social meaning requires edges defined by observable interactions and, crucially, by interaction valence: whether encounters are affiliative or agonistic [6, 20].

Advances in computer vision now make such interaction-aware networks feasible. Deep-learning systems can detect, identify, and track livestock under variable lighting, crowding, and partial occlusion [1]. Pose-based approaches extend location tracking by resolving relative orientation, approach trajectories, body alignment, and contact dynamics, features that help distinguish social encounters with different behavioural meanings. Several automated pipelines, however, end with behaviour labels or event counts [21, 22]. Recent work has begun to derive animal social networks from vision-based interactions [23], but three challenges remain insufficiently connected: auditing interaction labels after transfer to the target farm environment, preserving valence during network construction, and quantifying how network topology, community assignments, and individual rankings differ after interactions are separated by valence. The key analytical step is therefore not simply to recognize more events, but to transform classified events into social structure without discarding their meaning.

Here, we develop and evaluate a valence-aware framework that connects pose-based interaction inference with social network analysis in the milking-station area of a commercial dairy farm. We first audit the transferred classifier against manually reviewed clips, then convert quality-controlled encounters into a weighted interaction network and separate this representation into affiliative and agonistic subnetworks. We compare their connectivity, cohesion, community structure, and individual positions to test a specific proposition: pooling interactions of opposing valence can obscure biologically distinct forms of social organization within the observed group. In this formulation, automated interaction classification is not the endpoint of behavioural monitoring but the basis for social phenotyping. Here, social phenotyping denotes the quantitative description of how behaviourally distinct encounters are distributed across animals and dyads within an observed setting; it is not a diagnosis of welfare state or a claim of stable individual traits. The framework moves precision livestock farming from isolated event detection toward a relational account of competition, affiliation, and group organization, while providing a foundation for longitudinal, multi-zone studies that link network change to independent measures of animal welfare.

## 2. Materials and Methods

### 2.1 Study design, setting, animals and ethics

This observational study used video collected in the pre-milking holding area of a commercial dairy farm in Sussex County, New Brunswick, Canada. The area was immediately upstream of a voluntary milking station, where cows approached the system, passed through a gate and queued for milking. It covered approximately one quarter of the barn and concentrated animal movement within a spatially restricted field of view (Figure 1). The analysis therefore represents interactions observed within one resource-access zone, not the complete social network of the herd. Thirty-six individually identified cows entered the camera field of view during the observation session. Animals differed in their presence within the zone; consequently, low observed interaction activity could reflect limited opportunity for observation, limited social activity, or both. No regrouping, resource restriction or experimental intervention was introduced. Cows followed routine farm management and milking procedures throughout data collection.

The observational study protocol was reviewed and approved by the Animal Care and Use Committee (ACUC) of Dalhousie University (protocol no. 2024-026). Written informed consent was obtained from the farm owner for participation in the study. Data collection was non-invasive and observational, and no alteration of housing, milking or routine farm management was introduced.

**2.2 Video acquisition and temporal segmentation**

A single overhead GoPro Hero 13 camera recorded continuously from 10:30 to 18:09 in ultrawide mode at 4K resolution and 60 frames per second, yielding 7 h 39 min of video (27,540 s). The camera was positioned to maximize coverage of the holding area while limiting occlusion during periods of crowding. The recording was divided into 11 consecutive, non-overlapping processing windows (Videos 1–11). These windows were used both to process the video in manageable units and to construct descriptive, within-session network snapshots. The full-session network was the primary analytical representation. Because window durations differed, interaction frequency was also expressed as events per minute. Window boundaries and durations are reported with the temporal results.

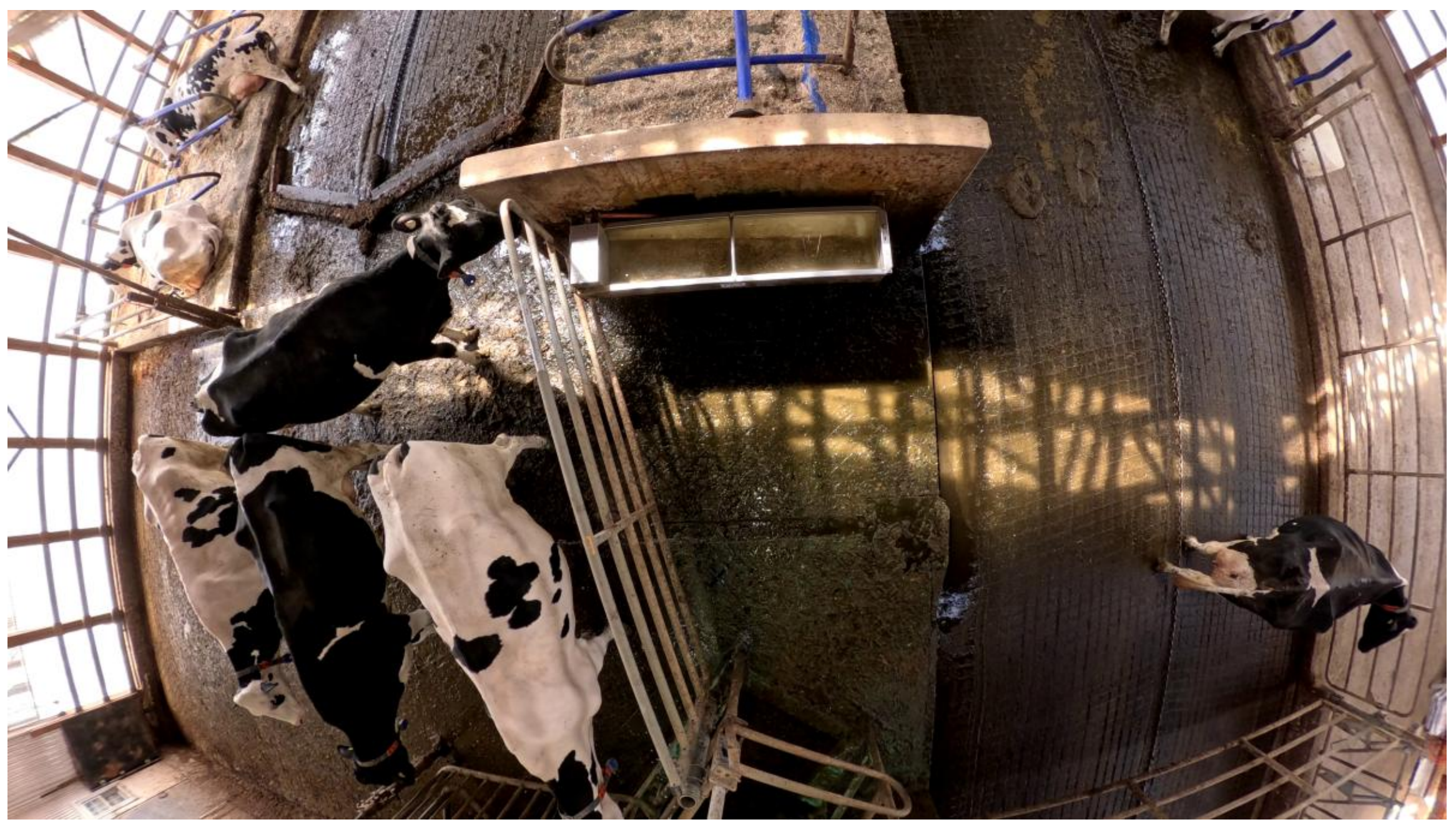

Figure 1. Study setting and observation zone. Overhead view of the commercial dairy farm's pre-milking holding area, showing the camera field of view used for video-based interaction analysis.

### 2.3 Automated interaction extraction

Dyadic events were extracted using the previously developed pose-based computer-vision pipeline for dairy-cattle social behaviour [24]. No new classifier was developed in the present study. Briefly, the pipeline detected cows in individual frames, linked detections through multi-object tracking, assigned persistent cow identities and estimated anatomical keypoints. Candidate dyads were generated from the animals' relative position, orientation, movement and persistence in proximity. Paired keypoint trajectories were then used to classify each candidate event as licking or grooming, displacement or headbutting. The model architecture, training procedure and event-inference framework are described in [24].

The present analysis began with the identity-resolved event records produced by this pipeline and converted them into valence-specific networks. It therefore evaluates the biological information retained when automated events are represented relationally, rather than proposing a new detection or classification model (Figure 2).

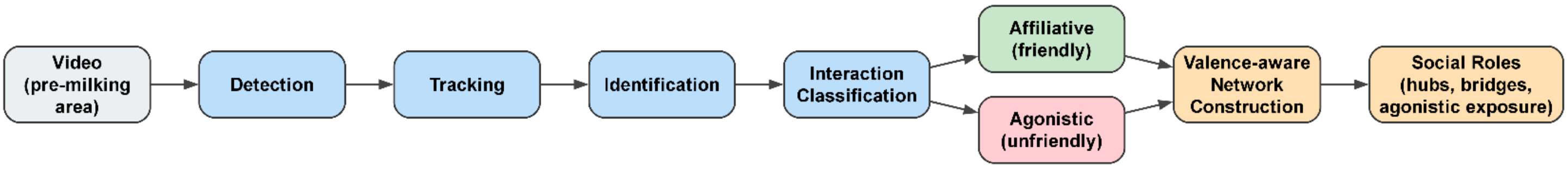

Figure 2. Computational workflow for constructing valence-aware social networks from video. Cows were detected, tracked and assigned persistent identities; pose trajectories were then used to identify and classify dyadic events. After quality control and production-domain auditing, automated event predictions were aggregated by dyad into weighted, undirected combined, affiliative and agonistic networks. Cumulative interaction duration defined edge weight.

### 2.4 Data quality and duration reconstruction

Records were excluded when either cow could not be identified reliably, track continuity was insufficient, or the exported event record was incomplete or malformed. Uncertain identities were not reassigned. Candidate, excluded and retained event totals are reported in the Results. An event began when both cows were detected. A detection gap of no more than 2 s was bridged within the same event to accommodate brief occlusion and was included in the reconstructed event interval; the first longer gap terminated the event. Sensitivity to this rule was examined by repeating duration reconstruction with gap thresholds from 2 to 10 s.

Two duration summaries were retained because simultaneous dyadic events affect them differently. For each cow, non-overlapping interaction time was calculated from the union of all event intervals involving that animal, so concurrent events with several partners contributed only once. Node strength was calculated as the sum of the durations of all incident dyadic edges; concurrent events with different partners therefore contributed separately to each relationship. Non-overlapping time describes elapsed involvement within the observed zone, whereas strength describes cumulative duration-weighted connectivity. Strength was not interpreted as causal social influence.

### 2.5 Behavioural classes and production-domain audit

Classifier outputs were mapped to affiliative and agonistic layers before graph construction (Table 1). Licking and grooming were treated as affiliative behaviours [9, 20]. Displacement and headbutting were treated as agonistic behaviours [7]. Here, agonistic denotes a competitive encounter and does not, by itself, indicate abnormal aggression or poor welfare. The deployed classifier produced affiliative or agonistic predictions. During manual auditing, a third label, no clear interaction, was available for clips that showed co-presence or tracking overlap without an unambiguous targeted social act. This label identified interaction-gate false positives; it was not treated as a third network layer. The audit quantified the performance of the automated dataset rather than manually relabelling the complete recording.

Table 1. Operational mapping of event labels to the analytical categories used in this study. Licking or grooming was treated as affiliative, displacement and headbutting as agonistic, and no clear interaction was used only during manual auditing.

| Event label | Analytical category | Operational description |
|---|---|---|
| Licking or grooming | Affiliative | Allogrooming or gentle muzzle contact directed towards another cow |
| Displacement | Agonistic | One cow causes another to yield its position or move away |
| Headbutting | Agonistic | Direct head-to-head or head-to-body contact |
| No clear interaction | Manual audit only | Co-presence or tracking overlap without an unambiguous affiliative or agonistic act |

Because the classifier had been developed primarily in a less crowded setting [24], its transfer to the commercial holding area was audited before the network outputs were interpreted. A predicted-class-stratified sample of 200 clips was selected across the 11 processing windows, with equal numbers of predicted affiliative and agonistic events. Two clips could not be evaluated, leaving 198 clips. Each evaluable clip was manually assigned one of three labels: affiliative, agonistic or no clear interaction.

Audit agreement was calculated as the proportion of all 198 clips for which an automated valence prediction matched the manual affiliative or agonistic label; clips labelled no clear interaction were counted as disagreements. A 95% Wilson confidence interval was calculated for this proportion. Precision, recall and F1 were calculated separately for the two predicted valence classes, with no-clear clips contributing false-positive predictions, and macro-F1 was the unweighted mean of the two class-specific F1 values. Conditional valence agreement was also calculated among clips manually judged to contain a clear affiliative or agonistic event. Because sampling began with pipeline-detected candidates and was stratified by predicted class, the audit estimates label agreement conditional on detection; it does not estimate missed-event frequency or end-to-end sensitivity across the continuous video.

### 2.6 Network construction

Networks were constructed in NetworkX using the quality-controlled event records. Each node represented an identified cow. Multiple events involving the same pair were collapsed into one edge, with cumulative event duration as the edge weight. Three full-session representations were generated: a combined network containing all retained predictions, an affiliative network containing licking or grooming events, and an agonistic network containing displacement or headbutting events. Edges were undirected because the annotation scheme encoded event type and valence but did not distinguish initiator from recipient. The graphs therefore describe whether, for how long and with what predicted valence a dyad interacted; they do not encode the direction of displacement, aggression or social influence.

The combined graph included every cow contributing at least one retained event. Each valence-specific graph included cows contributing at least one event of that valence; cows without an event in a layer were omitted from that layer's node set. Density and path-based comparisons between layers are therefore conditional on the animals participating in each valence-specific network. Edge weights were accumulated durations within the observation zone and were not treated as stable, context-independent measures of social preference.

### 2.7 Network metrics

The observed networks were described at network and node levels using standard graph-theoretic measures [13]. Unweighted topology was used for density, diameter, mean shortest-path length, global transitivity, degree assortativity, degree and betweenness centrality. Cumulative dyadic duration was used for strength, weighted local clustering, eigenvector centrality, PageRank and modularity. Betweenness was computed separately for each graph on its unweighted topology; the affiliative-network estimate was used for the primary interpretation of topological brokerage because subgroup structure was defined from affiliative ties. Duration was not converted directly to path length because a larger duration represented greater accumulated interaction, not greater network distance.

For the full-session layer comparisons, path-based metrics were calculated among the connected set of participating nodes in each graph. PageRank is reported by its statistical name rather than as a direct measure of biological influence. Table 2 gives the operational definition and edge treatment for each measure.

Table 2. Network measures used to characterize herd-level topology and individual structural position. All networks were undirected. Where indicated, edge weight was the cumulative duration of retained interactions within each dyad.

| Metric | Analytical level | Edge treatment | Operational definition |
|---|---|---|---|
| Density | Network | Unweighted | Proportion of all possible cow pairs connected by at least one retained interaction |
| Diameter | Network | Unweighted | Greatest number of edges in the shortest path between any two cows in the connected network |
| Mean shortest-path length | Network | Unweighted | Mean number of edges along the shortest paths connecting all pairs of cows |
| Global transitivity | Network | Unweighted | Proportion of connected triplets that are closed to form triangles |
| Mean weighted local clustering | Network | Duration-weighted | Mean, across cows, of the extent to which each cow's interaction partners were also interconnected, weighted by dyadic interaction duration |
| Degree assortativity | Network | Unweighted | Correlation between the degrees of cows joined by an edge; negative values indicate disassortative mixing |
| Weighted modularity ($Q$) | Network | Duration-weighted | Excess interaction weight within detected communities relative to that expected under a strength-preserving null model |
| Degree | Node | Unweighted | Number of distinct cows with which the focal cow had at least one retained interaction |
| Strength | Node | Duration-weighted | Total interaction duration accumulated by the focal cow across all its partners |
| Betweenness centrality | Node | Unweighted | Normalized proportion of shortest paths between other cow pairs that pass through the focal cow |
| Eigenvector centrality | Node | Duration-weighted | Centrality proportional to the summed centrality of neighbouring cows, scaled by dyadic interaction duration |
| PageRank | Node | Duration-weighted | Stationary probability of visiting the focal cow in a duration-weighted random walk with damping |

### 2.8 Community detection

Network modules were estimated by greedy modularity optimization using the Clauset-Newman-Moore algorithm and cumulative interaction duration as the edge weight. The algorithm was applied separately to the combined, affiliative and agonistic networks. The affiliative partition was used as the primary representation of subgroup structure because affiliative ties more directly encode positive association, whereas agonistic encounters primarily encode competition [6, 20]. The combined and agonistic partitions were retained for descriptive comparison. In this single-session analysis, a community denotes an algorithmically identified module with greater within-module than between-module connectivity under the modularity criterion. It should not be interpreted as evidence of a stable or persistent social group beyond the observed setting and period.

### 2.9 Within-session and dyadic analyses

A separate network was constructed for each of the 11 consecutive processing windows using the procedures described above. Nodes were cows contributing at least one retained event within that window. For each snapshot, the analysis summarized the number and rate of interactions, non-overlapping interaction duration, proportion of agonistic predictions, number of active cows and dyads, density, modular structure and node centralities. These snapshots characterize segment-specific variation within one recording session; they do not test long-term network stability or repeatability.

Dyadic structure was summarized by total event duration, affiliative duration, agonistic duration, event count and the number of processing windows in which each pair appeared. For dyad i,j, valence balance was calculated as $B_{ij} = T_{ij}^{aff} /(T_{ij}^{aff} + T_{ij}^{ago})$, where $T_{ij}^{aff}$ and $T_{ij}^{ago}$ denote cumulative affiliative and agonistic duration, respectively. Thus, $B_{ij} = 1$ indicates exclusively affiliative predicted interaction time and $B_{ij} = 0$ exclusively agonistic predicted interaction time. Recurrence across windows was reported descriptively; it was not interpreted as evidence of a persistent relationship beyond the observation session.

### 2.10 Analytical scope

All network comparisons were descriptive. No inferential test was used to establish that valence caused differences in topology, and differences between layers were not described as statistically significant. Individual cows were compared directly using degree, strength, betweenness, non-overlapping interaction time and agonistic-duration fraction. Narrative descriptions such as high connectivity or topological brokerage were based on these observed metric profiles; no prespecified welfare categories or diagnostic thresholds were applied.

The analysis was restricted to one holding area during one 7 h 39 min session. It did not adjust edge weights for individual time in view or dyadic co-presence opportunity, and it did not include independent production, health or welfare outcomes such as milk yield, somatic cell count, lameness, body condition or veterinary records. The networks therefore describe observed social involvement, valence and organization within this setting. They do not measure herd-wide social preference, stable individual traits or animal welfare state.

## 3. Results

### 3.1 Interaction yield and production-domain performance

The 7 h 39 min recording yielded 1,414 candidate interactions. Quality control removed 231 records with unresolved identities, insufficient track continuity, or incomplete event data, leaving 1,183 retained events involving 36 cows and 177 dyads (Table 3). Varying the bridged detection-gap threshold from 2 to 10 s retained the node and dyad sets and did not alter the leading individual rankings. Agonistic predictions accounted for 72.4% of events and 76.0% of cumulative interaction duration. These proportions characterize interactions detected within the pre-milking holding area; they do not quantify aggression or welfare state.

Table 3. Retained event counts and descriptive metrics for the combined undirected interaction network over the 7 h 39 min session. Mean local clustering and modularity were weighted by cumulative dyadic interaction duration; the remaining topological measures were unweighted. Agonistic percentages refer to automated predictions.

| Metric | Value |
|---|---|
| Observation duration | 7 h 39 min (27,540 s) |
| Cows contributing at least one retained event | 36 |
| Candidate events | 1,414 |
| Retained events | 1,183 |
| Excluded events | 231 |
| Unique retained dyads (edges) | 177 |
| Network density | 0.281 |
| Mean degree | 9.83 |
| Mean shortest-path length | 1.88 |
| Network diameter | 4 |
| Global transitivity | 0.513 |
| Mean duration-weighted local clustering | 0.503 |
| Degree assortativity | −0.170 |
| Duration-weighted modularity ($Q$) | 0.277 |
| Detected modules | 3 |
| Predicted agonistic events | 856 (72.4%) |
| Predicted agonistic dyadic duration | 76.0% |

Of the 200 clips selected for the production-domain audit, 198 were evaluable. Automated and manual valence labels agreed for 164 clips (82.8%; 95% Wilson confidence interval, 77.0–87.4%), with a macro-F1 of 0.872 (Figure 3). The 34 disagreements comprised 20 clips manually judged to contain no clear interaction and 14 clips in which affiliative and agonistic labels were reversed. Thus, interaction-gate false positives were more frequent than valence reversals in this audit. Among the 178 clips manually confirmed to contain an affiliative or agonistic interaction, conditional valence agreement was 92.1%. Because clips were sampled from pipeline-detected candidates, these estimates describe label agreement conditional on detection and do not measure missed interactions in the continuous video.

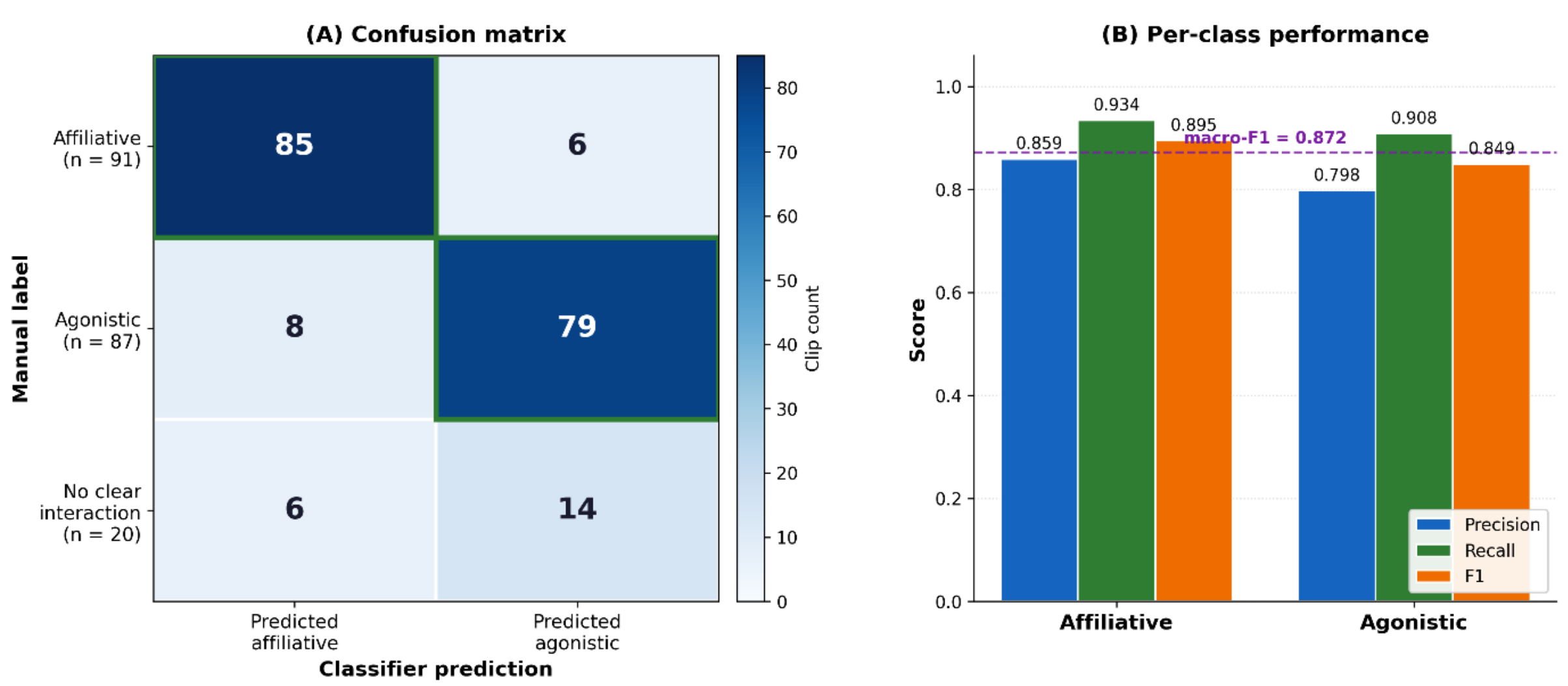

Figure 3. Production-domain audit of the interaction classifier in the pre-milking holding area (n = 198 clips). (A) Confusion matrix comparing manual labels with automated predictions; outlined cells denote concordant affiliative and agonistic classifications. (B) Class-specific precision, recall, and F1 (macro-F1 = 0.872). Overall agreement was 82.8% (164 of 198 clips; 95% Wilson confidence interval, 77.0–87.4%). Among the 178 clips containing a manually confirmed affiliative or agonistic interaction, conditional valence agreement was 92.1%. Twenty of the 34 disagreements were interaction-gate false positives and 14 were valence reversals.

### 3.2 Combined-network topology and individual position

The combined network contained all 36 cows in a single connected component (Figure 4). Its 177 edges represented 28.1% of the 630 possible dyads, giving a mean degree of 9.83. Network distances were short (mean shortest-path length, 1.88; diameter, 4), and triadic closure was substantial (global transitivity, 0.513; mean duration-weighted local clustering, 0.503). Degree assortativity was mildly negative (−0.170), indicating that higher-degree cows tended to connect with lower-degree cows. Weighted community detection identified three modules in the combined network (Q = 0.277). These descriptors establish compact and clustered topology within the observed interaction network, but no random-network comparison was performed to test small-world organization.

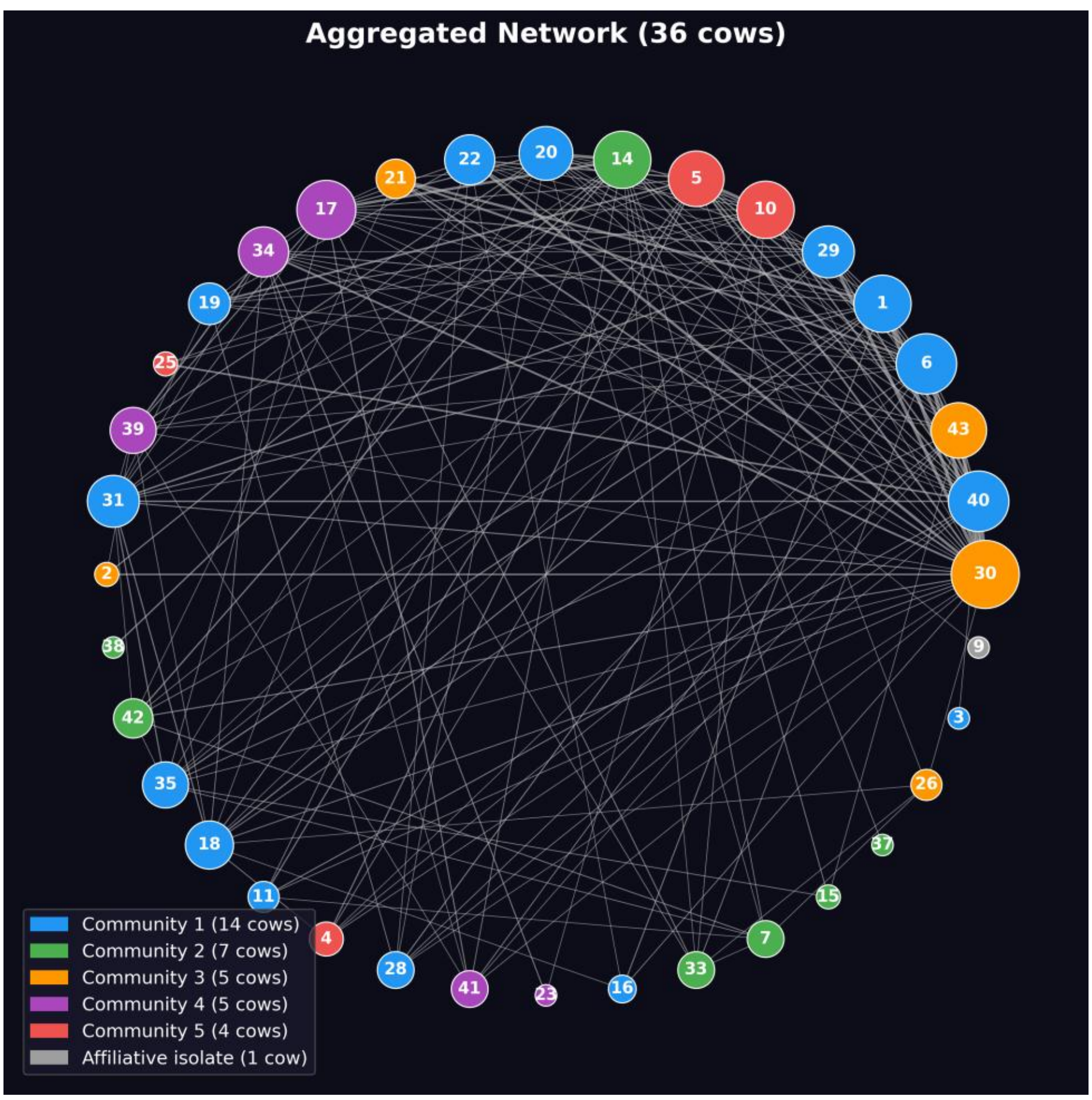

Figure 4. Combined undirected interaction network for the 36 observed cows. Node size represents degree, edge width represents cumulative dyadic interaction duration, and node colour represents

membership in the duration-weighted affiliative partition. Cow 9 had no retained affiliative event and is shown in grey. The colours therefore project the five-module affiliative partition onto the combined network; community detection on the combined network itself identified three modules. Node positions support visualization and do not represent physical locations in the barn.

Individual participation was strongly heterogeneous (Table 4). Cow 30 interacted with 24 of the 35 possible partners and accumulated 4,782 s of non-overlapping interaction time, equivalent to 17.4% of the recording. It also had the highest strength in the combined network. Cows 40, 6, 1, 14, 10, 17, 43, 20, and 5 each interacted with 15–19 partners, although their agonistic-duration fractions ranged from 58.6% to 94.0%. High activity therefore did not imply a uniform interaction profile.

Connectivity and brokerage also identified different animals. Cow 30 had the greatest degree, whereas Cow 14 had the highest betweenness in the affiliative network (0.170), despite ranking fifth by non-overlapping interaction time. Cow 30 ranked second in affiliative betweenness (0.156). At the other extreme, Cows 9, 3, 37, and 38 each interacted with one observed partner and appeared in only one processing window. Their low participation describes this camera field and recording session; it cannot establish herd-wide social isolation because observation opportunity was not standardized across cows.

Table 4. Per-cow descriptors for the ten cows with the greatest non-overlapping interaction time. Community membership and betweenness were calculated on the affiliative network; degree and non-overlapping interaction time describe the combined network. Agonistic time is the fraction of each cow's non-overlapping interaction time classified as agonistic.

| Cow | Affiliative community | Non-overlapping interaction time (s) | Recording time (%) | Degree | Agonistic time (%) | Affiliative betweenness centrality |
|---|---|---|---|---|---|---|
| 30 | 3 | 4,782 | 17.4 | 24 | 77.6 | 0.156 |
| 40 | 1 | 3,302 | 12.0 | 19 | 73.8 | 0.127 |
| 5 | 5 | 2,262 | 8.2 | 16 | 94.0 | 0.069 |
| 1 | 1 | 2,250 | 8.2 | 17 | 86.3 | 0.071 |
| 14 | 2 | 2,127 | 7.7 | 17 | 76.1 | 0.170 |
| 6 | 1 | 1,736 | 6.3 | 19 | 84.9 | 0.094 |
| 17 | 4 | 1,596 | 5.8 | 18 | 79.7 | 0.050 |
| 43 | 3 | 1,454 | 5.3 | 16 | 58.6 | 0.037 |
| 20 | 1 | 1,284 | 4.7 | 15 | 81.0 | 0.038 |
| 10 | 5 | 1,242 | 4.5 | 17 | 79.6 | 0.054 |

### 3.3 Affiliative communities and structural brokerage

The affiliative network contained 35 cows and separated into five modules of 14, 7, 5, 5, and 4 cows ($Q = 0.429$; Table 5). Cow 9 contributed no affiliative edge and was therefore absent from this layer. The largest module contained Cows 40, 1, 6, and 20. A seven-cow module containing Cow 14 had the lowest mean agonistic-duration fraction (46%), whereas the four-cow module containing Cow 5 had the highest (88%). Cow 30 belonged to a five-cow affiliative module that also contained Cow 2, whose observed interaction time was 99% affiliative. Cow 39, in a separate five-cow module, also had a predominantly affiliative profile (81.7%).

The affiliative partition was not a simple reproduction of the combined network. Cows 30 and 40 were assigned to different affiliative modules despite their prominent positions in the combined graph. Similarly, Cow 5 was separated from Cows 18, 19, and 22 in the affiliative partition, although all four showed high agonistic involvement. Separating predicted events by valence determined which retained edges contributed to each modular representation.

Table 5. Size and agonistic interaction profile of communities detected in the duration-weighted affiliative network.

| **Affiliative community** | **Cows, *n*** | **Mean cow-level agonistic time (%)** |
| --- | --- | --- |
| 1 | 14 | 79 |
| 2 | 7 | 46 |
| 3 | 5 | 57 |
| 4 | 5 | 59 |
| 5 | 4 | 88 |

*Note:* Five communities were identified by greedy modularity maximization ($Q$ = 0.429). Community labels are arbitrary and do not indicate rank. Mean agonistic time is the unweighted arithmetic mean of the cow-specific proportions of non-overlapping interaction time classified as agonistic. Of the 36 observed cows, 35 contributed at least one affiliative event and were included in this network. Cow 9 contributed no affiliative event and was therefore absent from the affiliative layer and its community partition. Agonistic time was used only to describe the communities after detection and did not contribute to the partition itself.

Betweenness further distinguished interaction volume from network position. Cow 14 had the highest affiliative betweenness (0.170), followed by Cow 30 (0.156). Cow 31 ranked eighteenth by non-overlapping interaction time but fifth by betweenness (0.074), while Cow 18 ranked fourteenth by time and eighth by betweenness (0.067). These contrasts show that extensive participation and placement on shortest affiliative paths were related but non-equivalent properties. Betweenness describes topological brokerage within the observed network; it does not demonstrate leadership, information transfer, or causal influence.

### 3.4 Affiliative and agonistic networks differ in cohesion and modularity

Separating events by predicted valence produced networks with descriptively different edge organization (Figure 5; Table 6). Of the 177 dyads in the combined network, 90 (50.8%) occurred in both valence layers, 21 occurred only in the affiliative layer, and 66 occurred only in the agonistic layer. The affiliative layer contained 111 edges and was less dense than the agonistic layer, which contained 156 edges (density, 0.187 versus 0.262). It also showed lower transitivity (0.325 versus 0.501) and lower mean duration-weighted local clustering (0.285 versus 0.452). In contrast, the affiliative layer was more modular, separating into five communities with Q = 0.429, whereas the agonistic layer separated into four communities with Q = 0.271. The combined network had the greatest density (0.281), because it contained the union of valence-specific edges, and an intermediate modularity (Q = 0.277).

Individual prominence was partly conserved across valence layers. Cow 30 led degree and strength in the combined, affiliative, and agonistic networks, although its degree differed between the affiliative and agonistic layers (18 versus 20). Cow 14 had the highest betweenness in all three representations (0.172 combined, 0.170 affiliative, and 0.155 agonistic). Thus, the valence-specific representations differed descriptively in cohesion and community partitioning without replacing the leading high-degree or high-betweenness cow. The principal contribution of valence separation is therefore not that every ranking changes, but that interactions with opposing behavioural meanings no longer contribute to the same edge structure.

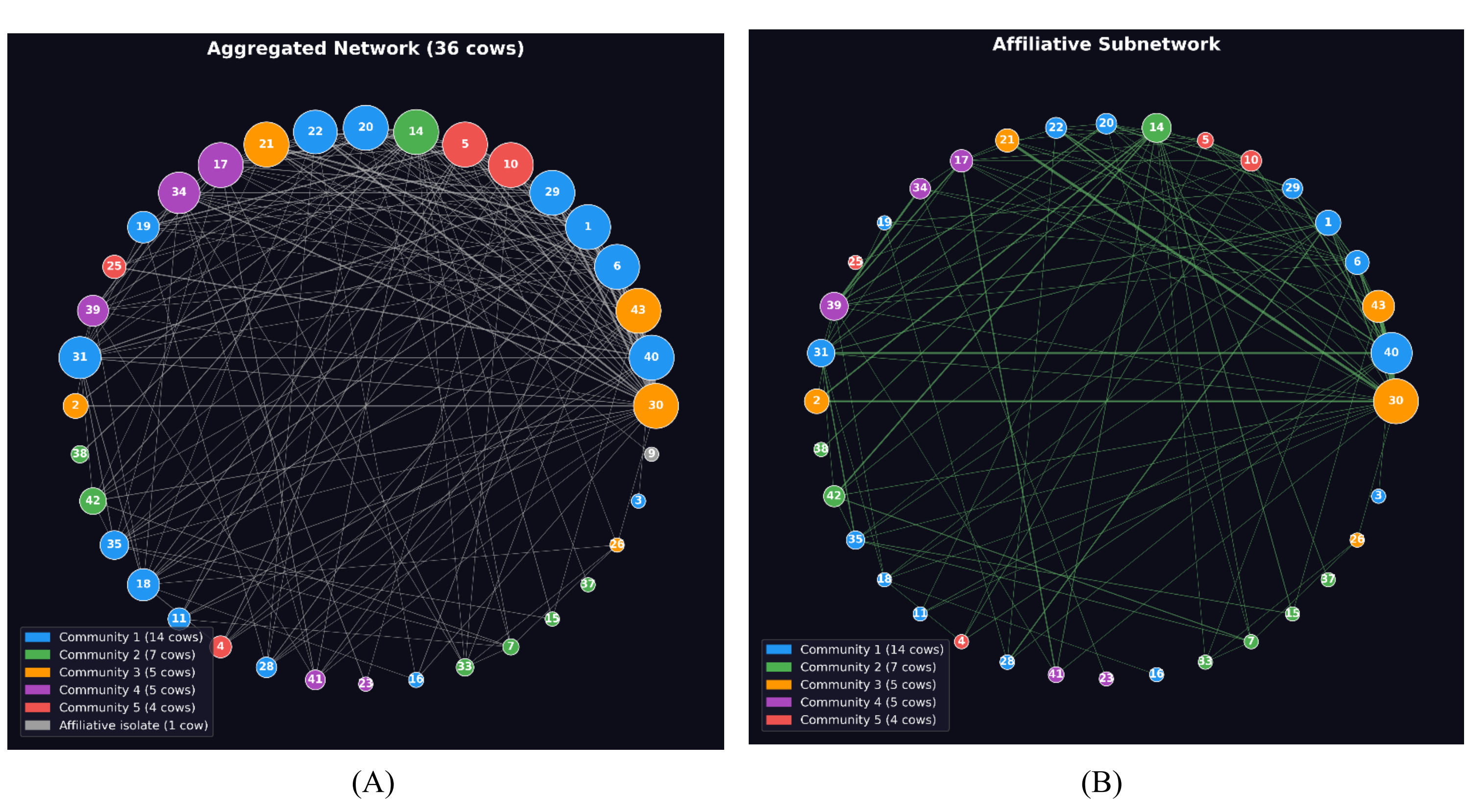

(A) (B)

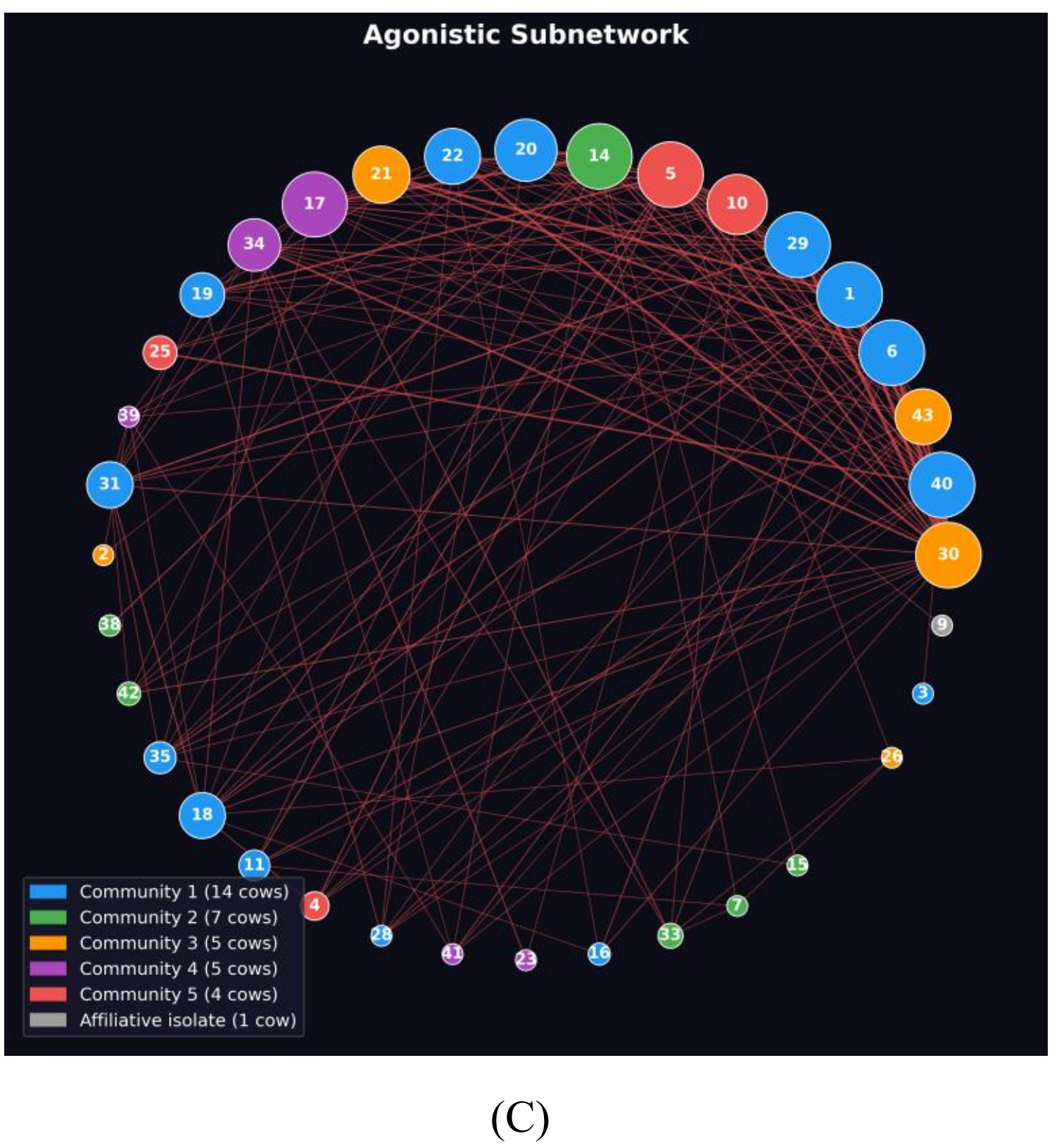

(C)

Figure 5. Combined and valence-specific interaction layers. The (A) combined, (B) affiliative, and (C) agonistic networks are shown at fixed node positions. Node colour encodes the full-session affiliative partition in every panel and does not represent independently detected communities in panels A and C. Node size represents within-layer strength, and edge width represents cumulative dyadic interaction duration. Grey, green, and red edges denote combined, affiliative, and agonistic ties, respectively. Cows without a retained event in a layer are omitted. Node positions support comparison and do not represent physical locations.

Table 6. Topology and leading node positions in the aggregated and valence-specific interaction networks.

| Metric | Aggregated | Affiliative | Agonistic |
|---|---|---|---|
| Nodes | 36 | 35 | 35 |
| Edges | 177 | 111 | 156 |
| Density | 0.281 | 0.187 | 0.262 |
| Global transitivity | 0.513 | 0.325 | 0.501 |
| Mean duration-weighted local clustering | 0.503 | 0.285 | 0.452 |
| Connected components | 1 | 1 | 1 |
| Duration-weighted modularity, $Q$ (communities) | 0.277 (3) | 0.429 (5) | 0.271 (4) |
| Highest-degree cow, ID (degree) | 30 (24) | 30 (18) | 30 (20) |
| Highest-strength cow, ID | 30 | 30 | 30 |
| Highest-betweenness cow, ID (centrality) | 14 (0.172) | 14 (0.170) | 14 (0.155) |

*Note:* All networks were undirected. The aggregated network included every retained interaction, whereas each valence-specific network included only cows contributing at least one retained event of the corresponding class. Node counts, and therefore the denominators used to calculate density and normalized centrality, differed between

layers. Edge weights represented cumulative dyadic interaction duration. Local clustering and modularity were duration-weighted, and node strength was defined from interaction duration. Density, transitivity, degree, connected components and betweenness centrality were calculated from binary network topology. Community counts are shown in parentheses after $Q$. Comparisons are descriptive; no inferential tests were performed.

### 3.5 Recurrent dyads and contrasting individual profiles

Dyadic interaction duration was concentrated among a limited set of pairs (Table 7). Cows 30 and 40 formed the highest-duration dyad, with 1,332 s of interaction across 60 events and five processing windows; 1,162 s was classified as agonistic. Other high-duration dyads were also predominantly agonistic. The 30–6, 1–30, 29–30, 10–5, 14–20, 22–30, 1–40, and 1–21 dyads each had an agonistic majority, and two of these pairs, 29–30 and 14–20, had no recorded affiliative duration. The 30–43 dyad differed: it recurred in six windows and had a mixed profile, with a modest affiliative majority (478 s affiliative and 398 s agonistic).

Recurrence across processing windows shows that the same dyads interacted repeatedly within the session. It does not distinguish a persistent social preference from repeated co-occurrence at the same resource bottleneck. The dyadic results nevertheless expose information lost through aggregation: pairs with similar total duration could differ markedly in predicted valence.

Table 7. Interaction profiles of the ten dyads with the greatest total interaction duration.

| Dyad | Total dyadic duration (s) | Affiliative duration (s) | Agonistic duration (s) | Affiliative balance, (B_{ij}) | Retained events, *n* | Event-positive windows, *n* | Duration-dominant valence |
|---|---|---|---|---|---|---|---|
| 30–40 | 1,332 | 170 | 1,162 | 0.128 | 60 | 5 | Agonistic |
| 30–43 | 875 | 478 | 398 | 0.546 | 16 | 6 | Affiliative |
| 30–6 | 871 | 86 | 785 | 0.099 | 28 | 2 | Agonistic |
| 1–30 | 765 | 88 | 677 | 0.115 | 45 | 2 | Agonistic |
| 29–30 | 731 | 0 | 731 | 0.000 | 8 | 4 | Agonistic |
| 5–10 | 708 | 131 | 577 | 0.185 | 35 | 2 | Agonistic |
| 14–20 | 671 | 0 | 671 | 0.000 | 4 | 2 | Agonistic |
| 22–30 | 598 | 108 | 490 | 0.181 | 29 | 2 | Agonistic |
| 1–40 | 574 | 7 | 568 | 0.012 | 10 | 2 | Agonistic |
| 1–21 | 536 | 1 | 536 | 0.002 | 3 | 2 | Agonistic |

*Note:* Total dyadic duration was calculated as the co-presence-capped union of all retained event intervals for each dyad, thereby preventing overlapping events from being counted more than once. Affiliative and agonistic durations were calculated separately. Affiliative balance was defined as $B_{ij} = T_{ij}^{\mathrm{aff}} / \left(T_{ij}^{\mathrm{aff}} + T_{ij}^{\mathrm{ago}}\right)$, where 0 denotes exclusively agonistic duration and 1 denotes exclusively affiliative duration. Duration-dominant valence identifies the class with the greater accumulated duration and does not imply an exclusive relationship type.

Because valence-specific intervals were rounded separately, their sum exceeds the reported union duration by 1 s for dyads 30–43, 1–40 and 1–21.

Degree, betweenness, non-overlapping interaction time, and agonistic-duration fraction described complementary aspects of individual participation (Table 8; Figure 6). Cow 30 combined the highest degree and greatest interaction time, whereas Cow 14 combined moderate interaction volume with the highest betweenness. Cows 5 and 1 had high agonistic-duration fractions, while Cows 9 and 3 had low observed degree. Cow 9 had no defined affiliative betweenness because it contributed no affiliative edge; Cow 3 had a value of zero. Cow 18 illustrates why these dimensions should not be converted into mutually exclusive roles: it combined relatively high betweenness with an agonistic-duration fraction of 96.5%.

The labels “hub”, “bridge”, “high agonistic involvement”, and “low observed participation” are used here as concise descriptions of observed metric profiles, not as prespecified biological categories. In particular, the undirected network cannot distinguish the initiator from the recipient of an agonistic event. A high agonistic-duration fraction therefore denotes involvement in agonistic interactions, not exposure as a recipient, dominance, or compromised welfare.

Table 8. Illustrative cows occupying contrasting positions across complementary dimensions of the observed interaction network.

| Feature highlighted | Cow ID | Combined-network degree | Affiliative betweenness centrality | Non-overlapping interaction time (s) | Agonistic share of interaction time (%) |
|---|---|---|---|---|---|
| High connectivity (hub-like) | 30 | 24 | 0.156 | 4,782 | 77.6 |
| High connectivity (hub-like) | 40 | 19 | 0.127 | 3,302 | 73.8 |
| High connectivity (hub-like) | 6 | 19 | 0.094 | 1,736 | 84.9 |
| High affiliative brokerage (bridge-like) | 14 | 17 | 0.170 | 2,127 | 76.1 |
| High affiliative brokerage (bridge-like) | 31 | 14 | 0.074 | 545 | 57.2 |
| High affiliative brokerage (bridge-like) | 18 | 12 | 0.067 | 659 | 96.5 |
| High agonistic-time proportion | 5 | 16 | 0.069 | 2,262 | 94.0 |
| High agonistic-time proportion | 1 | 17 | 0.071 | 2,250 | 86.3 |
| Low observed connectivity | 9 | 1 | NA | NA | NA |

| Feature highlighted | Cow ID | Combined-network degree | Affiliative betweenness centrality | Non-overlapping interaction time (s) | Agonistic share of interaction time (%) |
|---|---|---|---|---|---|
| Low observed connectivity | 3 | 1 | 0.000 | NA | NA |

*Note:* Profiles highlight contrasting network features and are neither mutually exclusive nor formal behavioural categories. Degree and non-overlapping interaction time were calculated from the aggregated network; normalized betweenness centrality was calculated from the affiliative network. Agonistic share denotes the proportion of each cow's non-overlapping interaction time classified as agonistic. Because interactions were represented as undirected, a high agonistic proportion indicates participation in agonistic encounters but does not identify the initiator or recipient and should not be interpreted as aggressiveness, dominance or welfare status. Cow 9 contributed no affiliative event and was therefore absent from the affiliative network, making its betweenness centrality undefined.

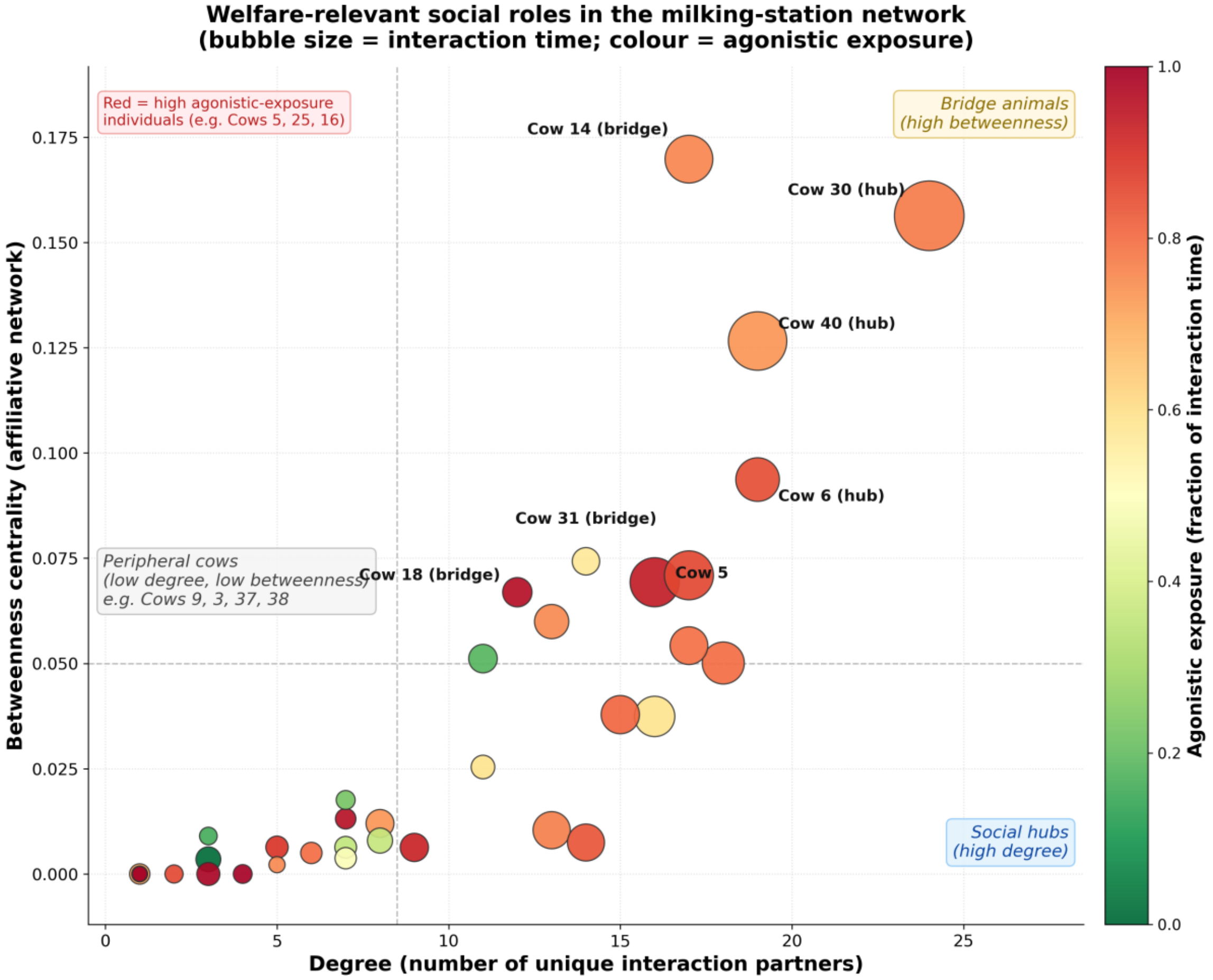

Figure 6. Individual variation in observed connectivity, affiliative brokerage, interaction time, and valence composition. Each cow is positioned by degree in the combined network and normalized betweenness in the affiliative network; bubble size represents non-overlapping interaction time and colour represents the predicted agonistic-duration fraction. Labels identify illustrative metric profiles rather than validated classes or welfare thresholds, and dashed guides are visual aids rather than analytical cut-offs. Cow 9 is displayed at zero betweenness only to indicate the absence of an affiliative

edge; its affiliative betweenness is undefined. Because the networks are undirected, agonistic involvement does not distinguish initiators from recipients.

### 3.6 Interaction activity varies within the recording session

Interaction activity differed markedly across the 11 processing windows (Figure 7; Table 9). Because window duration varied, rates provide the most direct comparison. The rate ranged from 0.1 to 7.9 events min−1, with 6–25 active cows and 3–66 active dyads per window. Video 1 (10:30–11:19) had the highest rate, with 388 events over 49 min (7.9 min−1), whereas Video 8 (15:47–16:26) had five events over 39 min (0.1 min−1) (Figure 8).

The proportion of events classified as agonistic ranged from 20.0% to 86.8%. This range should be interpreted alongside event counts: the 20.0% estimate in Video 8 was based on only five events, whereas the session-level estimate of 72.4% was based on 1,183. Cow 30 appeared in 10 of the 11 windows, showing that its prominence extended across most of the recording session. Other cows contributed to fewer windows or showed more localized peaks. These patterns establish within-session variation in observed participation; they do not demonstrate stable temporal roles beyond the recording day.

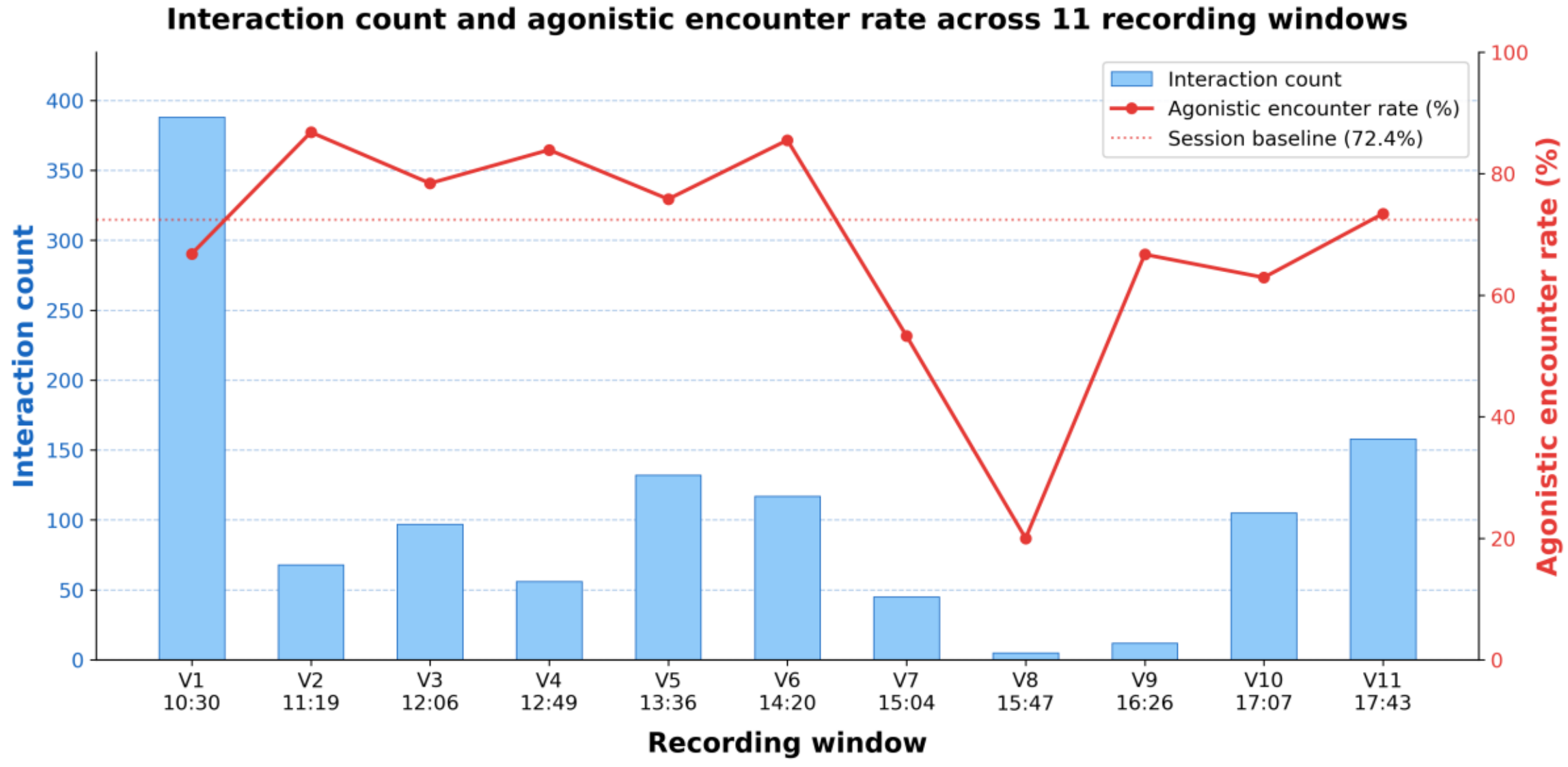

Figure 7. Retained interaction-event counts and predicted agonistic-event proportions across 11 consecutive processing windows. Bars show event counts and the line shows the proportion of retained events classified as agonistic; the horizontal reference denotes the event-weighted session proportion of 72.4%. Window durations differed from 26 to 49 min, and duration-adjusted event rates are reported in Table 9.

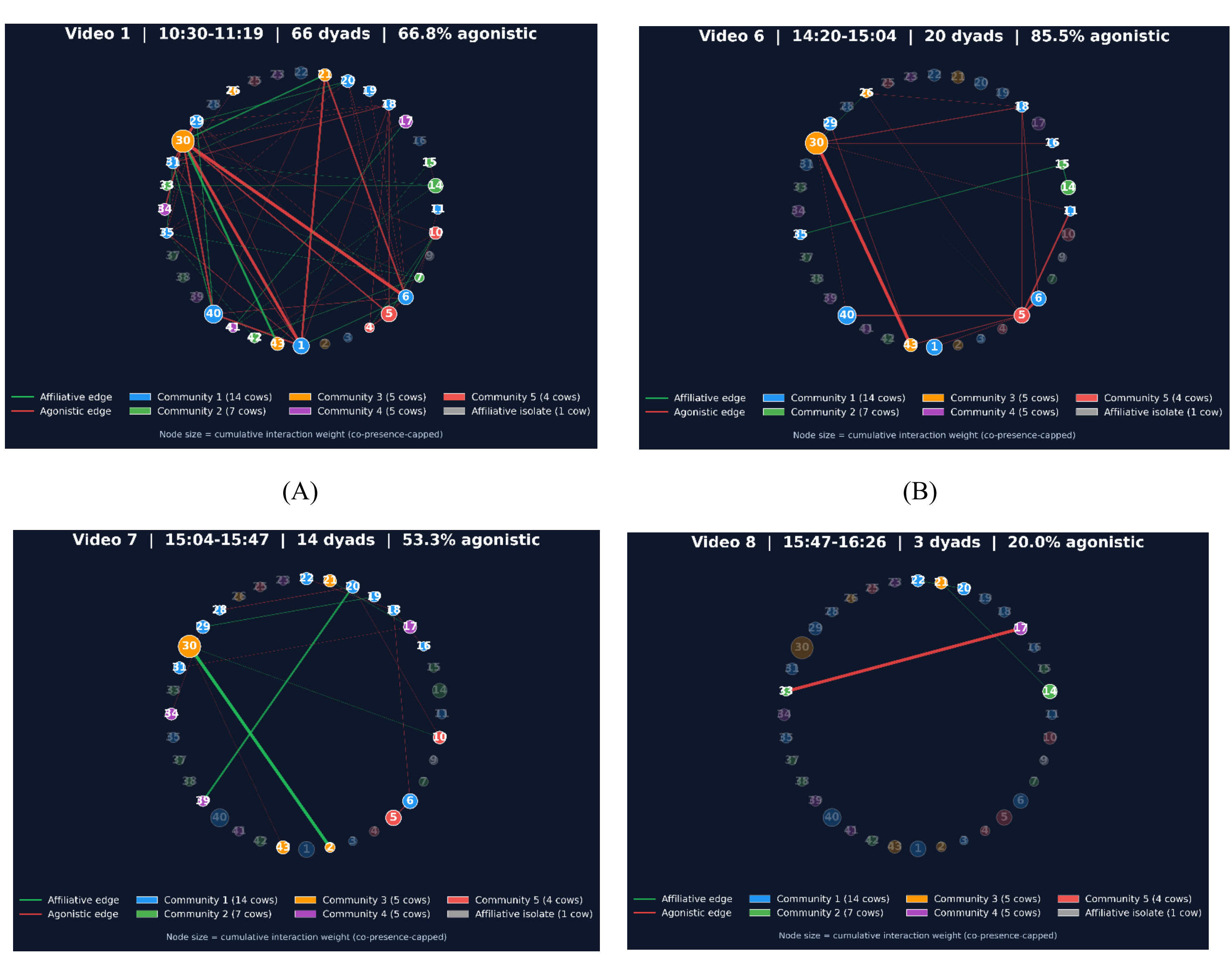

Figure 8. Within-session interaction-network snapshots. (A) Video 1, 10:30–11:19, 66 active dyads and 66.8% predicted agonistic events; (B) Video 6, 14:20–15:04, 20 active dyads and 85.5% predicted agonistic events; (C) Video 7, 15:04–15:47, 14 active dyads and 53.3% predicted agonistic events; and (D) Video 8, 15:47–16:26, three active dyads and 20.0% predicted agonistic events. Node positions and colours are fixed across panels; colours encode the full-session affiliative partition, node size represents within-window strength, and green and red edges denote predicted affiliative and agonistic interactions. Inactive cows are shown with reduced opacity. The layout does not represent physical location.

Table 9. Temporal distribution and predicted valence of retained interactions across 11 consecutive, non-overlapping recording windows.

| Recording window | Clock time | Retained interactions, *n* | Interaction rate (min$^{-1}$) | Active cows, *n* | Active dyads, *n* | Predicted agonistic interactions, *n* (%) |
|---|---|---|---|---|---|---|
| Video 1 | 10:30–11:19 | 388 | 7.9 | 25 | 66 | 259 (66.8) |
| Video 2 | 11:19–12:06 | 68 | 1.4 | 10 | 13 | 59 (86.8) |

| Recording window | Clock time | Retained interactions, *n* | Interaction rate ($min^{-1}$) | Active cows, *n* | Active dyads, *n* | Predicted agonistic interactions, *n* (%) |
|---|---|---|---|---|---|---|
| Video 3 | 12:06–12:49 | 97 | 2.3 | 18 | 29 | 76 (78.4) |
| Video 4 | 12:49–13:36 | 56 | 1.2 | 12 | 15 | 47 (83.9) |
| Video 5 | 13:36–14:20 | 132 | 3.0 | 19 | 33 | 100 (75.8) |
| Video 6 | 14:20–15:04 | 117 | 2.7 | 14 | 20 | 100 (85.5) |
| Video 7 | 15:04–15:47 | 45 | 1.0 | 18 | 14 | 24 (53.3) |
| Video 8 | 15:47–16:26 | 5 | 0.1 | 6 | 3 | 1 (20.0) |
| Video 9 | 16:26–17:07 | 12 | 0.3 | 11 | 8 | 8 (66.7) |
| Video 10 | 17:07–17:43 | 105 | 2.9 | 18 | 23 | 66 (62.9) |
| Video 11 | 17:43–18:09 | 158 | 6.1 | 20 | 38 | 116 (73.4) |
| Full session | 10:30–18:09 | 1,183 | 2.6 | 36 | 177 | 856 (72.4) |

*Note:* Interaction rates were calculated using the exact duration of each recording window. An active cow or dyad contributed at least one retained interaction during the corresponding window. Cows and dyads could recur across windows; therefore, the full-session row reports unique cows and dyads across the complete 459-min observation rather than sums of window-level counts. Agonistic classifications are model predictions. Window-specific percentages are descriptive, and values from sparsely populated windows, particularly Videos 8 and 9, should be interpreted alongside their event counts.

## 4. Discussion

Social networks are not neutral pictures of animal groups; their biological meaning depends on what is allowed to become an edge. In the pre-milking area studied here, pooling 1,183 retained affiliative and agonistic predictions produced a compact aggregated network. Separating those events by valence produced descriptively different edge sets, cohesion measures, and community partitions, while Cow 30 remained the highest-degree animal and Cow 14 retained the highest betweenness in each representation. The central result is not that valence altered the herd itself, but that aggregation altered what the network represented. An edge can combine encounters with opposing behavioural meanings, leaving a network that is topologically coherent but behaviourally ambiguous.

### 4.1 Valence resolves behaviourally distinct social structure

Interaction counts alone can conceal behavioural organization. The affiliative layer was sparser and less clustered than the agonistic layer, but more modular. It contained 111 rather than 156 edges, had a density of 0.187 rather than 0.262, and separated into five communities at Q = 0.429 rather than four at Q = 0.271. These comparisons were descriptive and were not tested against inferential or permutation-based null models. They therefore do not show that valence caused the topology of a stable herd network to change. They show that the same analytical choice, whether encounters of opposing meaning are pooled or retained separately, produces different representations of the observed social system.

Neither valence layer should be treated as the single "true" network. Affiliative and agonistic interactions describe different social processes, and earlier behavioural work has shown that their networks capture different aspects of cattle sociality [6, 9, 20]. The combined network answers who interacted; the affiliative layer describes how positive social contact was distributed; and the agonistic layer describes where competitive involvement occurred. These are complementary questions, not interchangeable measurements.

This distinction extends proximity-based approaches. Radio-frequency identification, real-time location systems, and related sensors efficiently measure co-location and opportunities for contact [14, 17, 18], but proximity cannot establish whether an encounter was affiliative, agonistic, or socially neutral. Earlier computer-vision work demonstrated that tracked cattle encounters can be assembled into social networks [23]. The present contribution is computational and behavioural: a production-domain audit of a pose-based classifier [24] is coupled to valence-specific network construction instead of ending with isolated event labels. Proximity and valence-aware monitoring are therefore best regarded as complementary. One establishes where contact is possible; the other begins to resolve what occurred.

The overlap between layers is as important as their differences. Of 177 retained dyads, 90 appeared in both valence networks. Valence is therefore a property of an encounter, not necessarily a permanent label for a pair. The 30–43 dyad illustrates this point: it recurred in six recording windows but included 478 s of predicted affiliative interaction and 398 s of predicted agonistic interaction. Calling this pair simply "affiliative" would conceal almost half of its observed behavioural content. Valence-aware analysis is most informative when it preserves such complexity rather than dividing cows into fixed categories such as friends and opponents.

**4.2 Competition and affiliation at the milking-area interface**

Agonistic predictions accounted for 72.4% of retained events and 76.0% of cumulative dyadic duration. The observed zone constrained movement around access to a shared milking facility, a context that may concentrate competitive encounters [8, 25]. However, no other zone, day, or herd was sampled, so the study cannot determine how much of the observed valence composition arose from facility layout, cow traffic, group composition, activity schedule, or classifier transfer. The values characterize the predicted composition of retained events in this resource-access zone during one session; they do not establish an aggressive herd temperament or poor welfare.

Affiliative organization nevertheless remained visible within this competition-rich setting. The higher modularity of the affiliative layer indicates that duration-weighted affiliative edges were more concentrated within the detected modules in this graph; it does not by itself demonstrate social preference. Familiarity, age, shared history, health, and regrouping can all shape cattle associations [3, 5, 12, 16], while allogrooming and other affiliative behaviours can identify preferred social partners [9, 20]. The present communities cannot, however, be equated with stable social bonds. They were inferred from one localized session, and repeated encounters may reflect common access schedules or co-occurrence at the milking system as well as social preference.

The high-duration dyads reinforce the need for this contextual restraint. Cows 30 and 40 accumulated 1,332 s of interaction, of which 1,162 s was predicted to be agonistic, whereas the recurrent 30–43 dyad remained mixed. Repeated agonistic encounters could reflect an established dominance relationship, closely matched competitors, or repeated convergence at the same bottleneck [4, 7]. Because the network was undirected, it cannot identify which cow initiated an event, which animal yielded, or whether the same behavioural asymmetry recurred. The data therefore describe agonistic involvement, not aggressiveness, dominance, victimization, or competitive "exposure."

Interaction activity also varied within the session. Event rates ranged from 0.1 to 7.9 min−1, and the predicted agonistic fraction ranged from 20.0% to 86.8%; several extreme percentages came from windows containing very few events. These patterns show that a session-level network averages over periods with different numbers of active cows, dyads, and detected encounters. They do not demonstrate temporal plasticity, or peaks in social stress. Longitudinal studies have shown that livestock networks can respond to disease or change in brokerage over time [11, 15], but comparable claims here require repeated observations with biologically defined time windows and adjustment for animal presence.

### 4.3 Network position is relational, not diagnostic

The node-level results show why social position cannot be reduced to a single score. Cow 30 led both degree and node strength and accumulated the greatest non-overlapping interaction time, but 77.6% of that time was predicted to be agonistic. Cow 14 had the highest betweenness despite ranking only fifth by non-overlapping interaction time. Breadth of connection, duration of participation, valence composition, and shortest-path position therefore described complementary aspects of the observed network.

These metrics should not be converted into fixed animal types. High degree does not by itself indicate sociability, and high betweenness does not demonstrate leadership, influence, information transfer, or causal importance to herd cohesion. Similarly, a high agonistic-duration fraction does not distinguish the initiator from the recipient of conflict. Low observed connectivity cannot establish social isolation because cows differed in their presence within the camera field and their opportunity to interact. Terms such as "hub-like," "brokerage-like," and "low observed participation" are therefore appropriate only as concise descriptions of this network and session.

### 4.4 From social phenotyping to welfare-informed farm decisions

The immediate practical value of valence-aware networks is not an automated welfare score. It is an additional behavioural layer for farm decision support. At zone level, repeated and presence-normalized measurements could identify where and when agonistic involvement becomes concentrated. At group level, networks recorded before and after regrouping or changes in stocking density, cow-flow procedures, and milking access could quantify disruption and recovery in affiliative structure and the concentration of competition [5]. At animal level, a sustained departure from a cow's own multi-zone baseline, such as declining affiliative connectivity, reduced observed participation, or repeated agonistic involvement, could prompt targeted observation of resource access, injury, lameness, or illness. Such alerts should guide inspection, not diagnose welfare or health.

This distinction is important because social change may be both a feature of the environment and a response to animal state. Disease challenge and health status have been associated with changes in livestock social networks [11, 12], and social unfamiliarity can alter integration and association patterns [3, 5]. Exploratory work in automatic milking systems further found that separation from an affinity partner was associated with approximately threefold greater day-to-day variability in milk production [26]. These findings make social-network measures plausible candidates for welfare- and health-oriented monitoring, but they do not validate the present metrics as outcomes or early-warning signals.

A credible commercial workflow would therefore be sequential. Automated video analysis would first identify persistent deviations in valence-specific event rates, recurrent dyads, community structure, or individual participation. Farm staff or veterinarians would then review the underlying clips and assess the animal directly. Network outputs would be interpreted alongside milk yield, feeding and rumination, locomotion and lameness scores, body condition, somatic cell count, treatment records, and direct welfare observations. This human-in-the-loop design is more useful and safer than applying universal centrality thresholds or assigning labels such as aggressive, isolated, or socially indispensable.

The same framework could support evaluation of housing and management. Before-and-after monitoring could test whether a redesigned holding area, altered gate timing, or a regrouping protocol reduces concentrated agonistic involvement without fragmenting affiliative associations. The present study did not compare interventions and cannot attribute competition to barn design. It establishes the measurement pathway needed to test those questions under commercial conditions.

### 4.5 Limitations and validation priorities

The findings describe 36 cows visible in one milking-associated zone during a single 7 h 39 min session. They cannot establish complete herd-wide relationships, day-to-day stability, persistent communities, or responses to regrouping. Animals also differed in time within the camera field, and edge weights were not normalized for individual visibility or dyadic opportunity to interact. Some apparent differences in connectivity may therefore reflect observation opportunity as well as social behaviour. Each valence layer included 35 cows, so layer-specific node sets provide an additional reason to treat topological comparisons as descriptive.

Valence labels were model predictions. In a predicted-class-stratified audit of 198 detected clips, automated and manual labels agreed in 82.8% of cases, with a macro-F1 of 0.872. This supports feasibility after transfer to a crowded production setting, but it is not an end-to-end estimate of performance. Sampling began with detected candidate events, so interactions missed in the continuous video were not measured; nor does the audit estimate the natural prevalence of either valence class. In addition, quality control removed 231 of 1,414 candidate events, making reproducible event filtering part of any operational workflow. Classification errors also propagate into edges, communities, and centrality, and an error involving a structurally prominent dyad may matter more than several peripheral errors. Future evaluation should therefore report not only event-level accuracy but also the stability of network conclusions under plausible detection and classification error.

Further limitations follow from the network design. Undirected edges precluded separation of agonistic initiation and receipt. Network layers were compared descriptively without inferential testing. No independent health, production, or welfare outcomes were available, so none of the metrics can be interpreted causally or assigned a welfare threshold.

Commercial and biological validation now require multi-camera, multi-zone, multi-day, and multi-herd observations. Directional coding should distinguish initiators from recipients; time in view and dyadic co-presence should be modelled as interaction opportunity; and parity, lactation stage, familiarity, health, and regrouping history should be included as covariates. Temporal or permutation-based null models are needed to test whether communities and repeated dyads exceed patterns expected from shared space use. Most importantly, prospective studies should determine whether network changes precede, accompany, or merely reflect independent measures of resource access, lameness, injury, physiological stress, milk production, somatic cell count, and veterinary treatment. Before-and-after interventions can then test whether changes to housing or management improve those outcomes.

The advance demonstrated here is methodological and behavioural, not diagnostic. Preserving predicted interaction valence converts automated events into testable descriptions of competition, affiliation, and localized group organization. With longitudinal validation against the animals' health, resource access, and welfare, those descriptions could become a practical evidence layer for managing socially complex dairy herds.

## 5. Conclusions

An interaction network can be mathematically well resolved yet biologically incomplete. Across 7 h 39 min of observation, 1,183 quality-controlled, classifier-labelled encounters among 36 cows and 177 dyads produced aggregated, affiliative and agonistic networks with different patterns of cohesion, community structure and individual position. The pooled network showed who interacted and for how long; the valence-specific layers revealed whether those ties reflected affiliation or competition. These forms of social exchange are not interchangeable. In social animals, the meaning of an edge is part of the biology.

The immediate contribution is not an automated welfare verdict, but an interpretable social-sensing layer for precision herd management. With longitudinal validation, valence-aware networks could flag persistent agonistic dyads, loss of affiliative ties, growing social peripherality or concentrations of conflict around particular times, locations and resources. These signals could help farm staff prioritize animal inspection, examine housing or resource bottlenecks, assess regrouping practices and select cows for closer behavioural, health or veterinary evaluation. Their greatest value will come from integration with clinical records, lameness and injury scores, feeding and resting access, milk production and validated animal-welfare measures. Network change should prompt investigation, not replace professional judgement.

This study was confined to one milking-station area, on one farm, during a single recording session. Valence was classifier-predicted, interactions were undirected, network comparisons were descriptive and no independent health, welfare or productivity outcomes were measured. The findings therefore do not establish stable social phenotypes, dominance, causation or the initiators and recipients of agonistic encounters. Multi-farm, multi-zone longitudinal studies must now test whether valence-specific network changes anticipate resource exclusion, disease, lameness or production loss, and whether management interventions modify those patterns. If validated, the practical value of computer vision will extend beyond detecting animals and behaviours to making the social environment of the herd measurable, interpretable and actionable.

**Data availability statement**

The curated video data underlying this study, together with sensitive farm-level metadata, are deposited in Zenodo under restricted access (https://doi.org/10.5281/zenodo.18234504). Access requests may be submitted through the Zenodo record or directed to the corresponding author and will require completion of a Data Access Agreement protecting farm confidentiality. Non-identifying metadata and supporting documentation are available through the Zenodo record. The computer-vision pipeline used to generate the interaction records is publicly available at https://github.com/mooanalytica/DairyCow-SNA.

**Ethics statement**

All experimental procedures were reviewed and approved by the Dalhousie University Animal Ethics Committee (Protocol No. 2024–026). The study was conducted in accordance with the local legislation and institutional requirements.

**Author contributions**

SP: Writing – original draft, Data curation, Methodology, Visualization, Formal analysis, Investigation, Conceptualization, Validation. SN: Project administration, Validation, Conceptualization, Resources, Writing – review & editing, Funding acquisition, Supervision.

**Funding**

This work was supported by the Natural Sciences and Engineering Research Council of Canada, the Nova Scotia Department of Agriculture, and the New Brunswick Department of Agriculture, Aquaculture and Fisheries.

**Acknowledgments**

The authors sincerely thank the Dairy Farmers of Nova Scotia and the Dairy Farmers of New Brunswick for generously providing access to their dairy farms, as well as for their valuable technical assistance and support throughout the multi-week data collection process. The authors also extend their gratitude to Jean Lynds, Operations Manager; Michael McConkey, Farm Manager; and Stewart Yuill, Animal Caretaker, for their unwavering support and assistance during data collection at the Ruminant Animal Centre, Dalhousie University.

**Conflict of interest**

The authors declare that the research was conducted in the absence of any commercial or financial relationships that could be construed as a potential conflict of interest. The author SN declared that they were an editorial board member of Frontiers at the time of submission. This had no impact on the peer review process and the final decision. SN is CEO of Agnovix Inc, which had no role in study design, data collection, analysis, or the decision to submit.